\documentclass[10pt,twocolumn,letterpaper]{article}

\usepackage{iccv}
\usepackage{times}
\usepackage{epsfig}
\usepackage{graphicx}
\usepackage{amsmath}
\usepackage{amssymb}

\usepackage{subcaption}
\usepackage{booktabs}
\usepackage{enumitem}
\usepackage{xcolor}
\usepackage{colortbl}
\usepackage[accsupp]{axessibility}  

\usepackage{cite}

\usepackage[colorlinks=true]{hyperref}

\iccvfinalcopy

\def\iccvPaperID{5230}

\newcommand{\satlas}{\mbox{\sc{SatlasPretrain}}}
\newcommand{\satnet}{\mbox{\sc{SatlasNet}}}

\definecolor{Gray}{gray}{0.92}
\newcolumntype{a}{>{\columncolor{Gray}}r}

\begin{document}

\title{SatlasPretrain: A Large-Scale Dataset for Remote Sensing Image Understanding}

\author{
Favyen Bastani \quad Piper Wolters \quad Ritwik Gupta \quad Joe Ferdinando \quad Aniruddha Kembhavi \\
Allen Institute for AI \\
{\tt\small {\{favyenb,piperw,ritwikg,joef,anik\}@allenai.org}}
}

\twocolumn[{%
\renewcommand\twocolumn[1][]{#1}%
\maketitle
\begin{center}
    \centering
    \captionsetup{type=figure}
    \includegraphics[width=\linewidth]{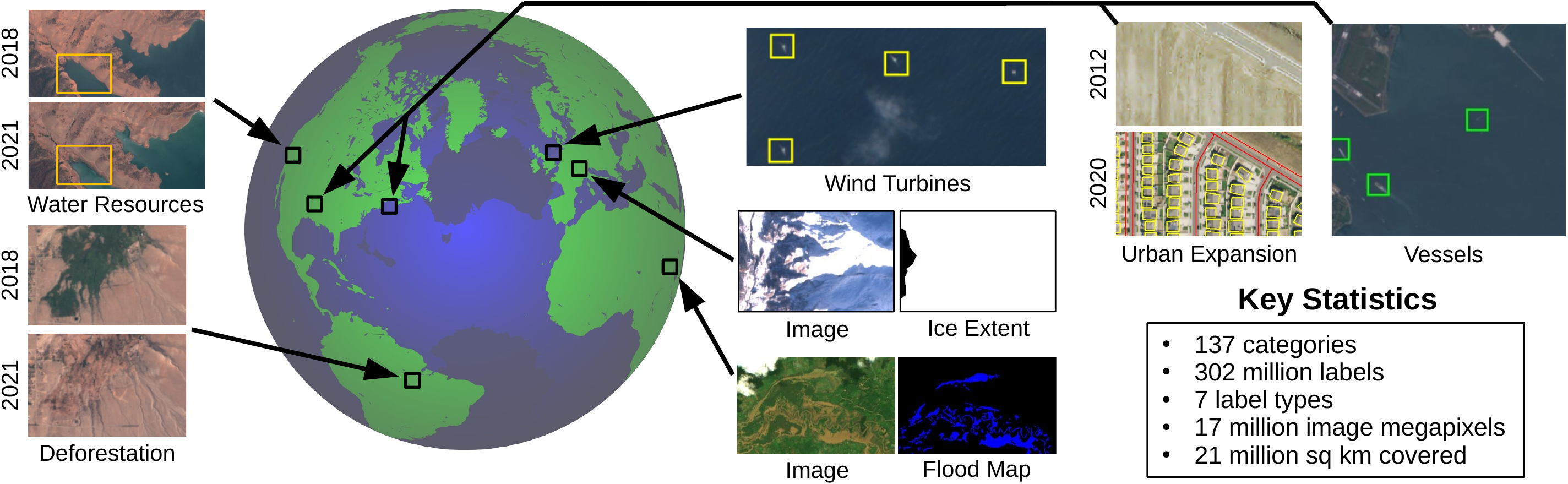}
    \captionof{figure}{\satlas\ is a large-scale remote sensing dataset. Its labels are relevant to many important planet monitoring applications, including water resource monitoring, tracking deforestation, detecting wind turbines for infrastructure mapping, tracking glacier loss, detecting floods, tracking urban expansion, and detecting vessels for tackling illegal fishing.}
    \label{fig:applications}
\end{center}%
}]

\begin{abstract}
Remote sensing images are useful for a wide variety of planet monitoring applications, from tracking deforestation to tackling illegal fishing. The Earth is extremely diverse---the amount of potential tasks in remote sensing images is massive, and the sizes of features range from several kilometers to just tens of centimeters. However, creating generalizable computer vision methods is a challenge in part due to the lack of a large-scale dataset that captures these diverse features for many tasks. In this paper, we present \satlas, a remote sensing dataset that is large in both breadth and scale, combining Sentinel-2 and NAIP images with 302M labels under 137 categories and seven label types. We evaluate eight baselines and a proposed method on \satlas, and find that there is substantial room for improvement in addressing research challenges specific to remote sensing, including processing image time series that consist of images from very different types of sensors, and taking advantage of long-range spatial context. Moreover, we find that pre-training on \satlas\ substantially improves performance on downstream tasks, increasing average accuracy by 18\% over ImageNet and 6\% over the next best baseline. The dataset, pre-trained model weights, and code are available at \url{https://satlas-pretrain.allen.ai/}.
\end{abstract}

\section{Introduction}
\label{sec:intro}

Satellite and aerial images provide a diverse range of information about the physical world.
In images of urban areas, we can identify unmapped roads and buildings and incorporate them into digital map datasets, as well as monitor urban expansion.
In images of industrial areas, we can catalogue solar farms and wind turbines to track the progress of renewable energy deployment.
In images of glaciers and forests, we can monitor slow natural changes like glacier loss and deforestation.
With the availability of global, regularly updated, and public domain sources of remote sensing images like the EU's Sentinel missions~\cite{sentinel}, we can monitor the Earth for all of these applications and more at a global-scale, on a monthly or even weekly basis.

Because the immense scale of the Earth makes global manual analysis of remote sensing images cost-prohibitive, automatic computer vision methods are crucial for unlocking their full potential.
Previous work has proposed applying computer vision for automatically inferring the positions of roads and buildings~\cite{roadtracer,sat2graph,dlinknet,batra2019improved,li2019topological,polyworld}; monitoring changes in land cover and land use such as deforestation and urban expansion~\cite{robinson2019large,robinson2020human}; predicting vessel positions and types to help tackle illegal fishing~\cite{xview3}; and tracking the progress and extent of natural disasters like floods, wildfires, and tornadoes~\cite{allison2016airborne,d2016bayesian,radhika2016application}.
However, in practice, most deployed applications continue to rely on manual or semi-automated rather than fully automated analysis of remote sensing images~\cite{analyzesatchallenge} for two reasons. First, accuracy remains a barrier even in major applications like road extraction~\cite{beyond_road_extraction}, making full automation impractical. Second, there is a long tail of remote sensing applications that require expert annotation but have few labeled examples (e.g., a recent New York Times study manually documented illegal airstrips in Brazil using satellite images~\cite{brazil_illegal_airstrips}).

We believe that the lack of a very-large-scale, multi-task remote sensing dataset is a major impediment for progress on automated methods for remote sensing tasks today.
First, state-of-the-art architectures such as ViT~\cite{vit} and CLIP~\cite{clip} require huge datasets to achieve peak performance.
However, existing remote sensing datasets for object detection, instance segmentation, and semantic segmentation like DOTA~\cite{dota}, iSAID~\cite{isaid}, and DeepGlobe~\cite{deepglobe} contain less than 10K images each, compared to the 328K images in COCO and millions used to train CLIP; the small size of these datasets means we cannot fully take advantage of recent architectures.
Second, existing remote sensing benchmarks are fragmented, with individual benchmarks for categories like roads~\cite{massroad}, vessels~\cite{xview3}, and crop types~\cite{pastisr}, but no benchmark spanning many categories. The lack of a large-scale, centralized, and accessible benchmark prevents transfer learning opportunities across tasks, and makes it difficult for computer vision researchers to engage in this domain.

We present \satlas, a large-scale dataset for improving remote sensing image understanding models. Our goal with \satlas\ is \emph{to label everything that is visible in a satellite image}. To this end, \satlas\ combines Sentinel-2 and NAIP images with 302M distinct labels under 137 diverse categories and 7 label types: the label types are \textbf{points} like wind turbines and water towers; \textbf{polygons} like buildings and airports; \textbf{polylines} like roads and rivers; \textbf{segmentation} and \textbf{regression} labels like land cover categories and bathymetry (water depth); \textbf{properties} of objects like the rotor diameter of a wind turbine; and \textbf{patch classification} labels like the presence of smoke in an image. Figure \ref{fig:applications} demonstrates the wide range of categories in \satlas, along with the diverse applications that they serve.

We find that the huge scale of \satlas\ enables pre-training to substantially improve downstream performance. We compare \satlas\ pre-training against pre-training on other datasets as well as self-supervised learning methods, and find that it improves average performance across seven downstream tasks by 18\% over ImageNet and 6\% over the next best baseline.
These results show that \satlas\ can readily improve accuracy on the numerous niche remote sensing tasks that require costly expert annotation.

Additionally, we believe that \satlas\ will encourage work on computer vision methods that tackle the unique research challenges in the remote sensing domain.
Compared to general-purpose computer vision methods, remote sensing models require specialized techniques such as accounting for long-range spatial context,
synthesizing information across images over time captured by diverse sensors like multispectral images and synthetic aperture radar (SAR),
and predicting objects that vary widely in size, from forests spanning many km$^2$ to street lamps.
We evaluate eight computer vision baselines on \satlas\ and find that no single existing method supports all the \satlas\ label types; instead, each baseline can only predict a subset of categories. Thus, inspired by recent work that integrate task-specific output heads~\cite{cho2021unifying,hu2021unit,gupta2022towards,kamath2022webly}, we develop a unified model called \satnet\ that incorporates seven such heads so that it can learn from every category in the dataset.
Compared to training separately on each label type, we find that jointly training \satnet\ on all categories and then fine-tuning on each label type improves average performance by 7.1\%, showing that \satnet\ is able to leverage transfer learning opportunities between label types.

In summary, our contributions are:
\begin{enumerate}[noitemsep]
    \item \satlas, a large-scale remote sensing dataset with 137 categories under seven label types.
    \item Demonstrating that pre-training on \satlas\ improves average performance on seven downstream datasets by 6\%.
    \item \satnet, a unified model that supports predictions for all label types in \satlas.
\end{enumerate}

We have released the dataset and code at \url{https://satlas-pretrain.allen.ai/}. We have also released model weights pre-trained on \satlas\, which can be fine-tuned for downstream tasks.

\section{Related Work} \label{sec:related}

\noindent
\textbf{Large-Scale Remote Sensing Datasets.} Several general-purpose remote sensing computer vision datasets have been released. Many of these focus on scene and patch classification: the UC Merced Land Use (UCM)~\cite{ucmerced_land_use} and BigEarthNet~\cite{bigearthnet} datasets involve land cover classification with 21 and 43 categories respectively, while the AID~\cite{aid}, Million-AID~\cite{millionaid}, RESISC45~\cite{resisc45}, and Functional Map of the World (FMoW)~\cite{fmow} datasets additionally include categories corresponding to manmade structures such as bridges and railway stations, with up to 63 categories. A few datasets focus on tasks other than scene classification. DOTA~\cite{dota} involves detecting objects in 18 categories ranging from helicopter to roundabout. iSAID~\cite{isaid} involves instance segmentation for 15 categories.

\begin{table}[]
    \centering
    \footnotesize
    \setlength\tabcolsep{3pt}
    \begin{tabular}{|l|r|r|r|r|r|}
        \hline
        & Types & Classes & Labels & Pixels & km$^2$ \\
        \hline
        \textbf{SatlasPretrain} & 7 & 137 & 302222K & 17003B & 21320K \\
        UCM~\cite{ucmerced_land_use} & 1 & 21 & 2K & 1B & 1K \\
        BigEarthNet~\cite{bigearthnet} & 1 & 43 & 1750K & 9B & 850K \\
        AID~\cite{aid} & 1 & 30 & 10K & 4B & 14K \\
        Million-AID~\cite{millionaid} & 1 & 51 & 37K & 4B & 18K \\
        RESISC45~\cite{resisc45} & 1 & 45 & 32K & 2B & 10K \\
        FMoW~\cite{fmow} & 1 & 63 & 417K & 437B & 1748K \\
        DOTA~\cite{dota} & 1 & 19 & 99K & 9B & 38K \\
        iSAID~\cite{isaid} & 1 & 15 & 355K & 9B & 38K \\
        \hline
    \end{tabular}
    \caption{Comparison of \satlas\ against existing remote sensing datasets (K=thousands, B=billions). Types is number of label types and km$^2$ is area covered.}
    \label{tab:dataset_comparison}
\end{table}

All of these datasets involve making predictions for a single label type, and most involve doing so from a single image. Thus, they are limited in three ways: the number of object categories, the diversity of labels, and the opportunities for approaches to learn to synthesize features across image time series. In contrast, \satlas\ incorporates 137 categories under seven label types (see full comparison in Table \ref{tab:dataset_comparison}), and provides image time series that methods can leverage to improve prediction accuracy.

A few domain-specific datasets extend beyond these limitations. xView3~\cite{xview3} involves predicting vessel positions (object detection) and attributes of those vessels such as vessel type and length (per-object classification and regression) in SAR images. PASTIS-R~\cite{pastisr} involves panoptic segmentation of crop types in crop fields using a time series of SAR and optical satellite images captured by the Sentinel-1 and Sentinel-2 constellations. IEEE Data Fusion datasets incorporate various aerial and satellite images for tasks like land cover segmentation~\cite{ieee_data_fusion}.

\smallskip
\noindent
\textbf{Self-Supervised and Multi-Task Learning for Remote Sensing.} Similar to our work, these approaches share the goal of improving accuracy on downstream applications with few labels. Several methods~\cite{seco,matter,scheibenreif2022contrastive,scheibenreif2022self,ssl4eo,bastani2021updating} incorporate temporal augmentations into a contrastive learning framework, where images of the same location captured at different times are encouraged to have closer representations than images of different locations. They show that the model improves downstream performance by learning invariance to transient differences between images of the same location, such as different lighting and nadir angle conditions as well as seasonal changes.
GPNA proposes combining self-supervised learning with supervised training on diverse tasks~\cite{gpna}.

\section{SatlasPretrain} \label{sec:satlas}

\begin{figure*}
    \centering
    \includegraphics[width=\linewidth]{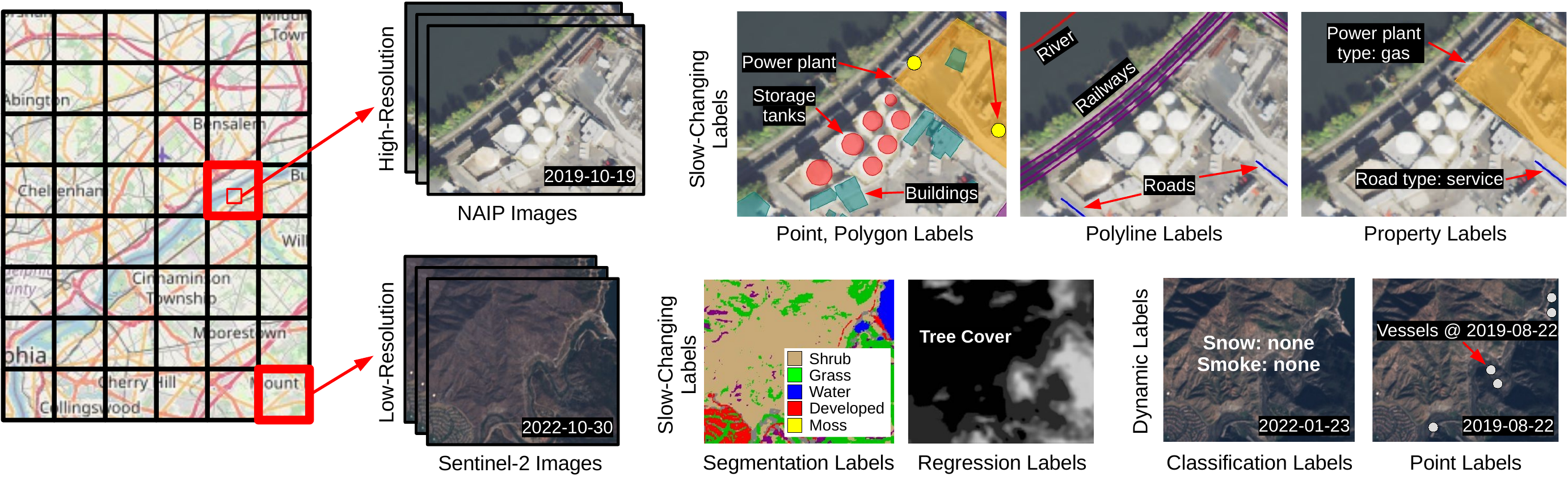}
    \caption{Overview of the \satlas\ dataset. \satlas\ consists of image time series and labels for 856K Web-Mercator tiles at zoom 13 (left). There are two image modes on which methods are trained and evaluated independently: high-resolution NAIP images (top) and low-resolution Sentinel-2 images (bottom). Labels may be slow-changing (corresponding to the most recent image at a tile) or dynamic (referencing a specific image and time).}
    \label{fig:dataset}
\end{figure*}

We present \satlas, a very-large-scale dataset for remote sensing image understanding that improves on existing remote sensing datasets in three key ways:

\begin{enumerate}[noitemsep, leftmargin=*]
    \item \textbf{Scale:} \satlas\ contains 40x more image pixels and 150x more labels than the largest existing dataset.
    \item \textbf{Label diversity:} Existing datasets in Table \ref{tab:dataset_comparison} have unimodal labels, e.g. only classification. \satlas\ labels span \emph{seven label types}; furthermore, they comprise 137 categories, 2x more than the largest existing dataset.
    \item \textbf{Spatio-temporal images and labels:} Rather than being tied to individual remote sensing images, our labels are associated with geographic coordinates (i.e., longitude-latitude positions) and time ranges. This enables methods to make predictions from multiple images across time, as well as leverage long-range spatial context from neighboring images. These features present new research challenges that, if solved, can greatly improve model performance.
\end{enumerate}

We first provide an overview of the structure of \satlas\ and detail the imagery that it contains below. We then describe the labels and how they were collected.

\subsection{Structure and Imagery} \label{sec:imagery}

\begin{figure}
    \centering
    \includegraphics[width=\linewidth]{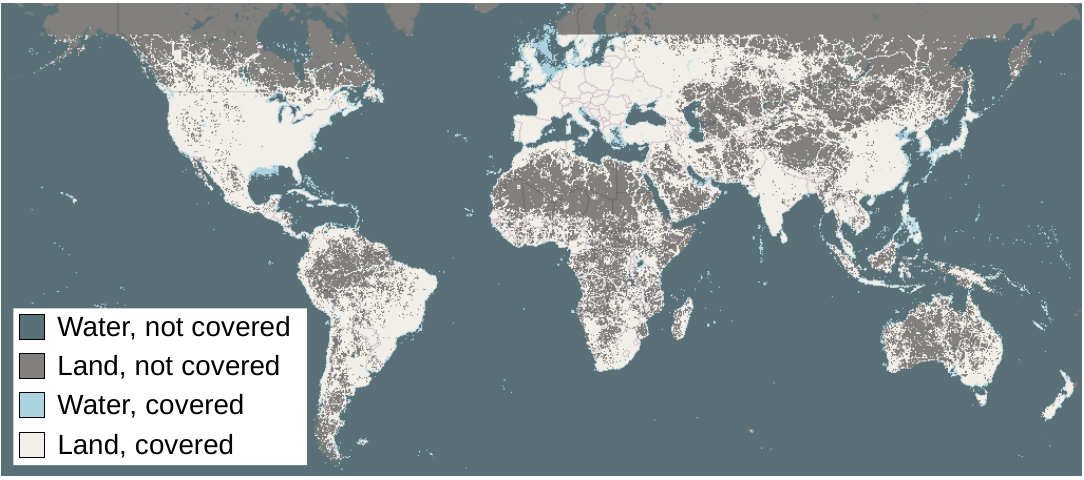}
    \caption{Geographic coverage of \satlas, with bright pixels indicating locations covered by images and labels in the dataset. \satlas\ spans all continents except Antarctica.}
    \label{fig:coverage}
\end{figure}

\satlas\ consists of 856K \emph{tiles}. These tiles correspond to Web-Mercator tiles at zoom level 13, i.e., the world is projected to a 2D plane and divided into a $2^{13} \times 2^{13}$ grid, with each tile corresponding to a grid cell. Thus, each \satlas\ tile covers a disjoint spatial region spanning up to 25 km$^2$. At each tile, \satlas\ includes (1) a time series of remote sensing images of the tile; and (2) labels drawn from the 137 \satlas\ categories. Figure \ref{fig:dataset} summarizes the dataset, and Figure \ref{fig:coverage} shows its global geographic coverage.

Existing datasets typically use either high-resolution imagery (0.5--2 m/pixel)~\cite{resisc45,millionaid,fmow,isaid} or low-resolution imagery (10 m/pixel)~\cite{bigearthnet,pastis}. Although high-resolution imagery enables higher prediction accuracy, low-resolution imagery is often employed in practical applications since it is available more frequently (weekly vs yearly) and broadly (globally vs in limited countries). Thus, in \satlas, we incorporate both low- and high-resolution images, which we will refer to as \emph{image modes}. We define separate train and test splits for each image mode, and compare methods over each mode independently.

In all 856K tiles (828K train and 28K test), we provide low-resolution 512x512 images. Specifically, we include 8--12 Sentinel-2 images captured during 2022; this enables methods to leverage multiple spatially aligned images of a location to improve prediction accuracy.
We also include historical 2016--2021 images that are relevant for dynamic labels like floods and ship positions. Sentinel-2 captures 10 m/pixel multispectral images; the European Space Agency (ESA) releases these images openly. Some categories are not visible in the low-resolution images, so for this mode we only evaluate methods on 122 of 137 categories.

In 46K tiles (45.5K train and 512 test), we provide high-resolution 8192x8192 images. We include 3--5 public domain 1 m/pixel aerial images from the US National Agriculture Imagery Program (NAIP) between 2011--2020. These images are only available in the US, so train and test tiles for the high-resolution mode are restricted to the US.

We download the images from ESA and USGS, and use GDAL~\cite{gdal} to process the images into Web-Mercator tiles.

The structure of \satlas\ enables methods to leverage both spatial and temporal context. Methods can make use of long-range spatial context from many neighboring tiles to improve the accuracy of predictions at a tile. Similarly, methods can learn to synthesize features across the image time series that we include at each tile in the dataset to improve prediction accuracy; for example, when predicting the crop type grown at a crop field, observations of the crop field at different stages of the agricultural cycle can provide different clues about the type of crop grown there.
In contrast, existing datasets (including all but FMoW in Table \ref{tab:dataset_comparison}) typically associate each label with a single image, and require methods to predict the label with that one image only.

\subsection{Labels}

\satlas\ labels span 137 categories, with seven label types (see examples in Figure \ref{fig:dataset}):
\begin{enumerate}[noitemsep, leftmargin=*]
    \item Semantic segmentation---e.g., predicting per-pixel land cover (water vs forest vs developed vs etc.).
    \item Regression---e.g., predicting per-pixel bathymetry (water depth) or percent tree cover.
    \item Points (object detection)---e.g., predicting wind turbines, oil wells, and vessels.
    \item Polygons (instance segmentation)---e.g., predicting buildings, dams, and aquafarms.
    \item Polylines---e.g., predicting roads, rivers, and railways.
    \item Properties of points, polygons, and polylines---e.g., the rotor diameter of a wind turbine.
    \item Classification---e.g., whether an image exhibits negligible, low, or high wildfire smoke density.
\end{enumerate}

\noindent
Most categories represent slow-changing objects like roads or wind turbines. During dataset creation, we aim for labels under these categories to correspond to the most recent image available at each tile. Thus, during inference, if these objects change over the image time series available at a tile, the model predictions should reflect the last image in the time series. A few categories represent dynamic objects like vessels and floods. For labels in these categories, in addition to specifying the object position, the label specifies the timestamp of the image that it corresponds to. During inference, for dynamic categories, the model should make a separate set of predictions for each image in the time series.

We derive \satlas\ labels from seven sources: new annotation by domain experts, new annotation by Amazon Mechanical Turk (AMT) workers, and processing five existing datasets---OpenStreetMap~\cite{openstreetmap}, NOAA lidar scans, WorldCover~\cite{worldcover}, Microsoft Buildings~\cite{msbuildings}, and C2S~\cite{c2sms}.

Each category is annotated (valid) in only a subset of tiles. Thus, in some tiles, a given category may be invalid, meaning that there is no ground truth for the category in that tile. In other tiles, a category may be valid but have zero labels, meaning that there are no instances of that category in the tile.
In supplementary Section A.1, for each category, we detail the number of tiles where the category is valid, the number of tiles where the category has at least one label, and the number of labels under that category; we also detail the category's label type and data source.

Labels in \satlas\ are relevant to numerous planet and environmental monitoring applications, which we discuss in supplementary Section A.2.

We summarize the data collection process for each of the data sources below.

\smallskip
\noindent
\textbf{Expert Annotation.} Two domain experts annotated 12 categories: off-shore wind turbines, off-shore platforms, vessels, 6 tree cover categories (e.g. low vs high), and 3 snow presence categories (none, partial, or full). To facilitate this process, we built a dedicated annotation tool called Siv that is customizable for individual categories. For example, when annotating marine objects, we found that displaying images of the same marine location at different times was crucial for accurately distinguishing vessels from fixed infrastructure (generally, a vessel will only appear in one of the images, while wind turbines and platforms appear in all images); thus, for these categories, we ensured the domain experts could press the arrow keys in Siv to toggle between different spatially aligned images of the same tile. Similarly, for tree cover, we found that consulting external sources like Google Maps and OpenStreetMap helped improve accuracy in cases where tree cover was not clear in NAIP or Sentinel-2 images; thus, when annotating tree cover, we included links in Siv to these external sources.

\smallskip
\noindent
\textbf{AMT.} AMT workers annotated 9 categories: coastal land, coastal water, fire retardant drops, areas burned by wildfires, airplanes, rooftop solar panels, and 3 smoke presence categories (none, low, or high). We reused the Siv annotation tool for AMT annotation, incorporating additional per-category customizations as needed (which we detail in supplementary Section A.3.1).

To maximize annotation quality, for each category, we first selected AMT workers through a qualification task: domain experts annotated between 100--400 tiles, and we asked each candidate AMT worker to annotate the same tiles; we only asked workers whose labels corresponded closely with expert labels to continue with further annotation.
We also conducted majority voting over multiple workers; we decided the number of workers needed per tile on a per-category basis (see Section A.3.2), by first having one worker annotate each tile, and then analyzing the label quality. For example, we found that airplanes were unambiguous enough that a single worker sufficed, while we had three workers label each tile for areas burned by wildfires.

\begin{figure*}
\centering
    \includegraphics[width=\linewidth]{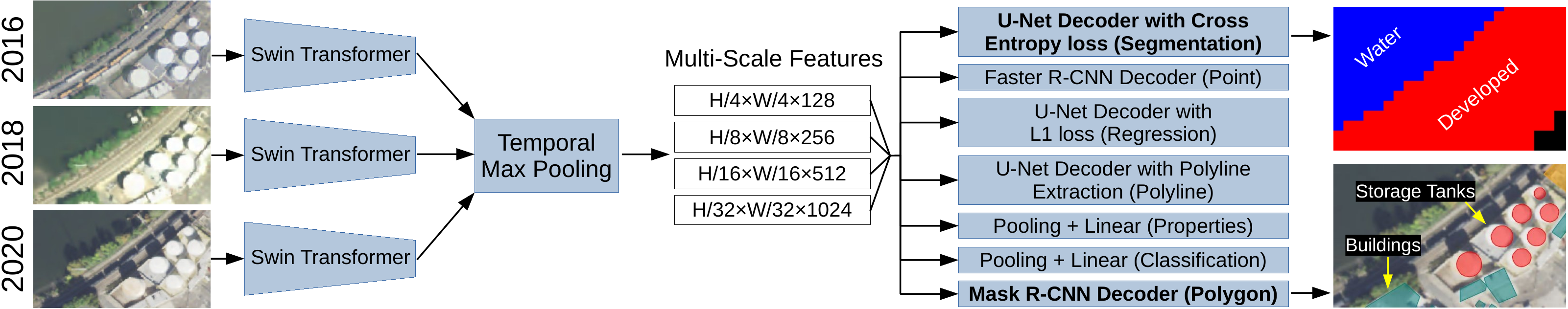}
\caption{Model architecture of \satnet. A separate head is used to predict outputs for each label type. We visualize example outputs from two such heads (segmentation and polygon).}
\label{fig:satnet}
\end{figure*}

\smallskip
\noindent
\textbf{OpenStreetMap (OSM).} OSM is a collaborative map dataset built through edits made by contributing users. Objects in OSM span a wide range of categories, from roads to power substations. We obtained OSM data as an OSM PBF file on 9 July 2022 from Geofabrik, and processed it using the Go osmpbf library to extract 101 categories.

Recall is a key issue for labels derived from OSM. From initial qualitative analysis, we consistently observed that OSM objects have high precision but variable recall: the vast majority of objects were correct, but for some categories, many objects were visible in satellite imagery but not mapped in OSM. To mitigate this issue, we employed heuristics to automatically prune tiles that most likely had low recall, based on the number of labels and distinct categories in the tile. For instance, we found that tiles with many roads but no buildings were likely to have missing objects in other categories like silo or water tower. We detail these heuristics in supplementary Section A.4.

We found that these heuristics were sufficient to yield high-quality labels for most categories. However, we identified 13 remaining low-recall categories, including gas stations, helipads, and oil wells. From an analysis of 1300 tiles, we determined that recall was still at least 80\% for these categories, which we deemed sufficient for the training set: there are methods for learning from sparse labels, and large-scale training on noisy labels has produced models like CLIP that deliver state-of-the-art performance. However, we deemed that these 13 categories did not have sufficient recall for the test set. Thus, to ensure a highly accurate test set, for each of these 13 categories, we trained an initial model on OSM labels and tuned its confidence threshold for high-recall low-precision detection; we then hand-labeled its predictions to add missing labels to the test set. In Section A.4, we detail these categories and the number of missing labels identified in the test set.

\smallskip
\noindent
\textbf{NOAA Lidar Scans.} NOAA coastal topobathy maps derived from lidar scans contain elevation data for land and depth data for water. We download 5,868 such maps from various NOAA surveys, and process them to derive per-pixel depth and elevation labels for 5,123 SatlasPretrain tiles.

\smallskip
\noindent
\textbf{WorldCover.} WorldCover~\cite{worldcover} is a global land cover map developed by the European Space Agency. We process the map to derive 11 land cover and land use categories, ranging from barren land to developed areas.

\smallskip
\noindent
\textbf{Microsoft Buildings.} We process 70 GeoJSON files from various Microsoft Buildings datasets~\cite{msbuildings} to derive building polygons in \satlas. The data is released under ODbL.

\smallskip
\noindent
\textbf{C2S}. C2S~\cite{c2sms} consists of flood and cloud labels in Sentinel-2 images, released under CC-BY-4.0. We warp the labels to Web-Mercator and include them in \satlas. We also download and process the Sentinel-2 images that correspond exactly to the ones used in C2S, so that they share the same processing as other Sentinel-2 images in \satlas.

\smallskip
\noindent
Balancing the scale of labels with label quality was a key consideration in managing new annotation and selecting existing data sources to process. As we developed the dataset, we conducted iterative analyses to evaluate the precision and recall of labels that we collected, and used this information to improve later annotation and refine data source processing. In supplementary Section A.5, we include an analysis of incorrect and missing labels under every category in the final dataset; we find that 116/137 categories have {\textgreater}99\% precision, 15 have 95-99\% precision, 4 have 90-95\% precision, and 2 have 80-90\% precision.

\section{SatlasNet} \label{sec:method}

Off-the-shelf computer vision models cannot handle all the label types in \satlas, e.g., while Mask2Former~\cite{mask2former} can simultaneously perform semantic and instance segmentation, it is not designed to predict properties of polygons or classify images.
This prevents these models from leveraging the full set of transfer learning opportunities present in \satlas; for example, detecting building polygons is likely useful for segmenting images for land cover and land use, since land use includes a human-developed category.
We develop a unified model, \satnet, that is capable of learning from all seven label types.

Figure \ref{fig:satnet} shows a schematic of our model.
\satnet\ is inspired by recent work that employ task-specific output heads~\cite{cho2021unifying,hu2021unit,gupta2022towards}, as well as methods that synthesize features across remote sensing image time series~\cite{fmow,pastis}.
It inputs a time series of spatially aligned images, and processes each image (which may contain more than three bands) through a Swin Transformer~\cite{swin} backbone (Swin-Base), which outputs feature maps for each image at four scales.
We apply max temporal pooling at each scale to derive one set of multi-scale features.
We pass the features to seven output heads (one for each label type) to compute outputs.
For polylines, while specialized polyline extraction architectures have been shown to improve accuracy~\cite{roadtracer,sat2graph,vecroad}, we opt to employ the simpler segmentation approach~\cite{dlinknet} where we apply a UNet head to segment images for polyline categories, and post-process the segmentation probabilities with binary thresholding, morphological thinning, and line following and simplification~\cite{cheng2017automatic} to extract polylines.

\begin{table*}
    \centering
    \footnotesize
    \setlength\tabcolsep{4pt}
    \begin{tabular}{|l|r|r|r|r|r|r||r|r|r|r|r|r|r|}
        \hline
        & \multicolumn{6}{c||}{High-Resolution NAIP Images} & \multicolumn{7}{c|}{Low-Resolution Sentinel-2 Images} \\
        \cline{2-14}
        Method & Seg $\uparrow$ & Reg $\downarrow$ & ~Pt $\uparrow$~ & Pgon $\uparrow$ & Pline  $\uparrow$ & Prop $\uparrow$ & Seg $\uparrow$ & Reg $\downarrow$ & ~Pt $\uparrow$~ & Pgon $\uparrow$ & Pline  $\uparrow$ & Prop $\uparrow$ & Cls $\uparrow$ \\
        \hline
        PSPNet (ResNext-101)~\cite{pspnet} & 77.8 & 15.0 & - & - & 53.2 & - & 62.1 & 16.2 & - & - & 30.7 & - & - \\
        LinkNet (ResNext-101)~\cite{linknet} & 77.3 & 12.9 & - & - & 61.0 & - & 61.1 & 14.1 & - & - & 41.4 & - & - \\
        DeepLabv3 (ResNext-101)~\cite{deeplabv3} & 80.1 & 10.6 & - & - & 59.8 & - & 61.8 & 13.9 & - & - & 44.7 & - & - \\
        ResNet-50~\cite{resnet} & - & - & - & - & - & 87.6 & - & - & - & - & - & 70.3 & 97 \\
        ViT-Large~\cite{vit} & - & - & - & - & - & 78.1 & - & - & - & - & - & 66.9 & 99 \\
        Swin-Base~\cite{swin} & - & - & - & - & - & 87.1 & - & - & - & - & - & 69.4 & 99 \\
        Mask R-CNN (ResNet-50)~\cite{maskrcnn} & - & - & 27.6 & 30.4 & - & - & - & - & 22.0 & 12.3 & - & - & - \\
        Mask R-CNN (Swin-Base)~\cite{maskrcnn} & - & - & 30.4 & 31.5 & - & - & - & - & 25.6 & 15.2 & - & - & - \\
        ISTR~\cite{istr} & - & - & 2.0 & 4.9 & - & - & - & - & 1.2 & 1.4 & - & - & - \\
        \hline
        SatlasNet (single-image, per-type) & 79.4 & 8.3 & 28.0 & 30.4 & 61.5 & 86.6 & 64.8 & 9.3 & 25.7 & 14.8 & 42.5 & 67.5 & 99 \\
        SatlasNet (single-image, joint) & 74.5 & 7.4 & 28.0 & 31.1 & 60.9 & 87.3 & 55.8 & 10.6 & 22.0 & 10.3 & 45.5 & 73.8 & 99 \\
        SatlasNet (single-image, fine-tuned) & 79.8 & \textbf{7.2} & 32.3 & 33.0 & \textbf{62.4} & \textbf{89.5} & 65.3 & 9.0 & 27.4 & 16.3 & 45.9 & \textbf{80.0} & 99 \\
        SatlasNet (multi-image, per-type) & 79.4 & 8.2 & 25.8 & 27.5 & 59.2 & 77.3 & 67.2 & 10.5 & 31.9 & 19.0 & 48.1 & 67.1 & 99 \\
        SatlasNet (multi-image, joint) & 79.2 & 7.8 & 31.2 & 33.8 & 53.6 & 87.8 & 66.7 & 8.5 & 31.5 & 19.5 & 41.9 & 78.8 & 99 \\
        SatlasNet (multi-image, fine-tuned) & \textbf{81.0} & 7.6 & \textbf{33.2} & \textbf{34.1} & 61.1 & 89.2 & \textbf{69.7} & \textbf{7.8} & \textbf{32.0} & \textbf{20.2} & \textbf{50.4} & \textbf{80.0} & 99 \\
        \hline
    \end{tabular}
    \caption{Results on the \satlas\ test set for the high- and low-resolution image modes. We break down results by label type: segmentation (Seg), regression (Reg), points (Pt), polygons (Pgon), polylines (Pline), properties (Prop), and classification (Cls). We show absolute error for Reg (lower is better), and accuracy for the others (higher is better).}
\label{tab:results}
\end{table*}

\section{Evaluation} \label{sec:eval}

We first evaluate our method and eight classification, semantic segmentation, and instance segmentation baselines on the \satlas\ test split in Section \ref{sec:satlas_results}. We then evaluate performance on seven downstream tasks in Section \ref{sec:downstream}, comparing pre-training on \satlas\ to pre-training on other remote sensing datasets, as well as self-supervised learning techniques specialized for remote sensing.

\subsection{Results on SatlasPretrain} \label{sec:satlas_results}

\smallskip
\noindent
\textbf{Methods.} We compare \satnet\ against eight baselines on \satlas. We select baselines that are either standard models or models that provide state-of-the-art performance for subsets of label types in \satlas. None of the baselines are able to handle all seven \satlas\ label types. For property prediction and classification, we compare ResNet~\cite{resnet}, ViT~\cite{vit}, and Swin Transformer~\cite{swin}. For segmentation, regression, and polylines, we compare PSPNet~\cite{pspnet}, LinkNet~\cite{linknet}, and DeepLabv3~\cite{deeplabv3}. For points and polygons, we compare Mask R-CNN~\cite{maskrcnn} and ISTR~\cite{istr}.

We train three variants of \satnet:
\begin{itemize}[noitemsep,nolistsep]
    \item Per-type: train separately on each label type.
    \item Joint: jointly train across all categories.
    \item Fine-tuned: fine-tune the jointly trained parameters on each label type.
\end{itemize}

All baselines are fine-tuned on each label type (after joint training on the subset of label types they can handle), which provides the highest performance.

For each \satnet\ variant, we also evaluate in single-image and multi-image modes.
For all baselines and single-image \satnet, we sample training examples by either (a) sampling a tile, and pairing the most recent image at the tile with slow-changing labels (with dynamic and other invalid categories masked); or (b) sampling a tile and image, and pairing the image with corresponding dynamic labels.
For multi-image \satnet, we provide as input a time series of eight Sentinel-2 images for low-resolution mode or four NAIP images for high-resolution mode; for slow-changing labels, the images are ordered by timestamp, but for dynamic labels, we always order the sampled image at the end of the time series input.
In all cases, we sample examples based on the maximum inverse frequency of categories appearing in the example. We use RGB bands only here, but include results for single-image \satnet\ with nine Sentinel-2 bands in supplementary Section C.

Across all methods, we input 512x512 images during both training and inference; for high-resolution inference, since images covering the tile are 8K by 8K, we independently process 256 512x512 windows and merge the model outputs. We employ random cropping, horizontal and vertical flipping, and random resizing augmentations during training. We initialize models with ImageNet-pretrained weights. We use the Adam optimizer, and initialize the learning rate to $10^{-4}$, decaying via halving down to $10^{-6}$ upon plateaus in the training loss. We train with a batch size of 32 for 100K batches.

\smallskip
\noindent
\textbf{Metrics.} We use standard metrics for each label type: accuracy for classification, F1 score for segmentation, mean absolute error for regression, mAP accuracy for points and polygons, and GEO accuracy~\cite{geo} for polylines.
We compute metrics per-category, and report the average across categories under each label type.

\begin{figure*}
    \centering
    \includegraphics[width=\linewidth]{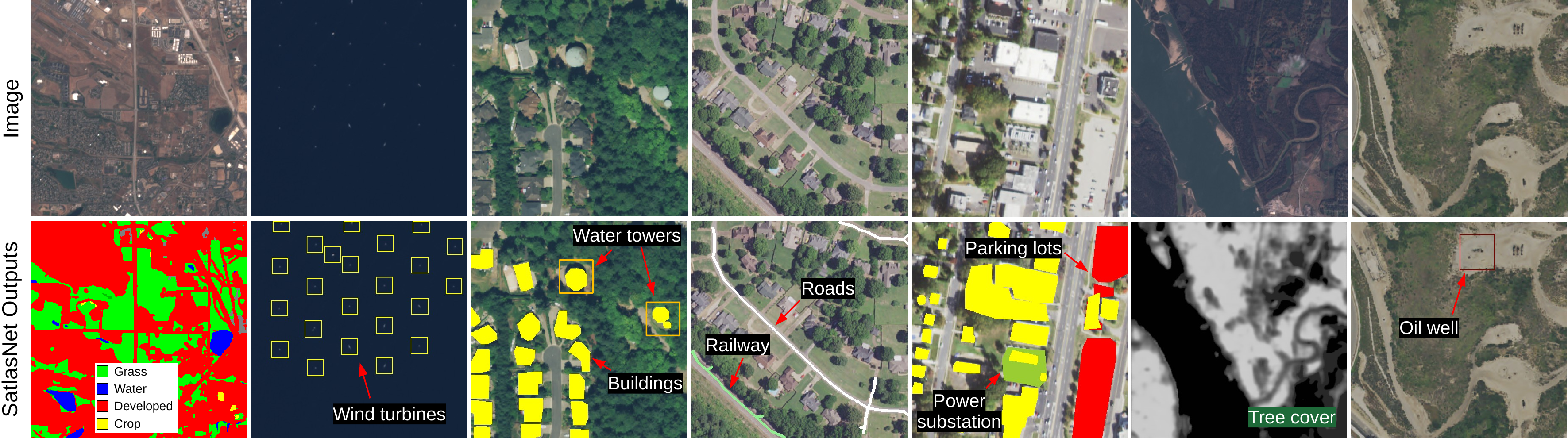}
    \caption{Qualitative results on \satlas. Rightmost: a failure case where \satnet\ detects only 1/5 oil wells.}
    \label{fig:qualitative}
\end{figure*}

\begin{table*}
    \centering
    \footnotesize
    \setlength\tabcolsep{5pt}
    \begin{tabular}{|l|a|r|a|r|a|r|a|r|a|r|a|r|a|r|a|r|}
        \hline
        & \multicolumn{2}{c|}{UCM} & \multicolumn{2}{c|}{RESISC45} & \multicolumn{2}{c|}{AID} & \multicolumn{2}{c|}{FMoW} & \multicolumn{2}{c|}{Mass Roads} & \multicolumn{2}{c|}{Mass Buildings} & \multicolumn{2}{c|}{Airbus Ships} & \multicolumn{2}{|c|}{\textbf{Average}}\\
        \cline{2-17}
        Method & 50 & All & 50 & All & 50 & All & 50 & All & 50 & All & 50 & All & 50 & All & 50 & All\\
        \hline
        Random Initialization & 0.26 & 0.86 & 0.15 & 0.77 & 0.18 & 0.68 & 0.03 & 0.17 & 0.69 & 0.80 & 0.68 & 0.77 & 0.31 & 0.53 & 0.33 & 0.65\\
        ImageNet~\cite{imagenet} & 0.35 & 0.92 & 0.17 & 0.95 & 0.20 & 0.81 & 0.03 & 0.21 & 0.77 & 0.85 & 0.78 & 0.83 & 0.37 & 0.65 & 0.38 & 0.75\\
        BigEarthNet~\cite{bigearthnet} & 0.35 & 0.95 & 0.20 & 0.94 & 0.23 & 0.78 & 0.03 & 0.27 & 0.78 & 0.85 & 0.81 & 0.85 & 0.40 & 0.68 & 0.40 & 0.76\\
        MillionAID~\cite{millionaid} & 0.72 & 0.97 & 0.30 & 0.96 & 0.30 & 0.82 & 0.04 & 0.35 & 0.78 & 0.84 & 0.82 & 0.85 & 0.46 & 0.67 & 0.49 & 0.78\\
        DOTA~\cite{dota} & 0.56 & \textbf{0.99} & 0.28 & 0.95 & 0.33 & 0.83 & 0.03 & 0.30 & \textbf{0.82} & 0.86 & 0.84 & 0.87 & \textbf{0.62} & 0.75 & 0.50 & 0.79\\
        iSAID~\cite{isaid} & 0.60 & 0.97 & 0.29 & 0.97 & 0.34 & 0.86 & 0.04 & 0.30 & \textbf{0.82} & 0.86 & 0.84 & 0.86 & 0.55 & 0.73 & 0.50 & 0.79\\
        MoCo~\cite{mocov2} & 0.14 & 0.14 & 0.07 & 0.09 & 0.05 & 0.12 & 0.02 & 0.03 & 0.56 & 0.69 & 0.62 & 0.63 & 0.01 & 0.21 & 0.21 & 0.27\\
        SeCo~\cite{seco} & 0.48 & 0.95 & 0.20 & 0.90 & 0.27 & 0.74 & 0.03 & 0.26 & 0.70 & 0.81 & 0.71 & 0.77 & 0.27 & 0.54 & 0.38 & 0.71\\
        \hline
        SatlasPretrain & \textbf{0.83} & \textbf{0.99} & \textbf{0.36} & \textbf{0.98} & \textbf{0.42} & \textbf{0.88} & \textbf{0.06} & \textbf{0.44} & \textbf{0.82} & \textbf{0.87} & \textbf{0.87} & \textbf{0.88} & 0.56 & \textbf{0.80} & \textbf{0.56} & \textbf{0.83}\\
        \hline
    \end{tabular}
    \caption{Results on seven downstream tasks when fine-tuned with 50 examples (50) or the entire downstream dataset (All). Accuracy is reported for UCM, RESISC45, and AID while F1 Score is reported for FMoW, Mass Roads, Mass Buildings, and Airbus Ships. \satlas\ pre-training improves average accuracy across the tasks by 6\% over the next best baseline.}
\label{tab:downstream}
\end{table*}

\smallskip
\noindent
\textbf{Results.} We show results on \satlas\ in Table \ref{tab:results}.
Across the seven label types, single-image \satnet\ matches or surpasses the performance of state-of-the-art, purpose-built baselines when trained per-type, validating its effectiveness as a unified model that can predict diverse remote sensing labels. Jointly training one set of \satnet\ parameters for all categories reduces average performance on several label types, but \satnet\ remains competitive in most cases; this training mode provides large efficiency gains since the backbone features need only be computed once for each image during inference, rather than once per label type. When fine-tuning \satnet\ on each label type using the parameters derived from joint training, it provides an average 7.1\% relative improvement across the label types and image modes over per-type training. This supports our hypothesis that there are transfer learning opportunities between the label types, validating the utility of a unified model for improving performance. Multi-image \satnet\ provides another 5.6\% relative improvement in average performance, showing that it is able to effectively synthesize information across image time series to produce better predictions; nevertheless, we believe that there is substantial room for further improvement in methods for processing remote sensing image time series.

We show qualitative results in Figure \ref{fig:qualitative}, with additional examples in supplementary Section E. We achieve high accuracy on several categories, such as wind turbines and water towers. However, for oil wells, one well is detected but several others are not. Similarly, for polyline features like roads and railways, the model produces short noisy segments, despite ample training data for these categories; we believe that incorporating and improving models that are tailored for specialized output types like polylines~\cite{sat2graph,vecroad} has the potential to improve accuracy.

\subsection{Downstream Performance} \label{sec:downstream}

We now evaluate accuracy on seven downstream tasks when pre-training on \satlas\ compared to pre-training on four existing remote sensing datasets, as well as two self-supervised learning methods. For each downstream task, we evaluate accuracy when training on just 50 examples and when training on the whole dataset, to focus on the challenge of improving performance on niche remote sensing applications that require expert annotation and thus have few labeled examples.

\smallskip
\noindent
\textbf{Methods.} We compare pre-training on high-resolution images in \satlas\ to pre-training on four existing remote sensing datasets: BigEarthNet~\cite{bigearthnet}, Million-AID~\cite{millionaid}, DOTA~\cite{dota}, and iSAID~\cite{isaid}.
We use \satnet\ in all cases, fine-tuning the pre-trained Swin backbone on each downstream dataset.

We also compare two self-supervised learning methods, Momentum Contrast v2 (MoCo)~\cite{mocov2} and Seasonal Contrast (SeCo)~\cite{seco}. The latter is a specialized method for remote sensing that leverages multiple image captures of the same location to learn invariance to seasonal changes. For MoCo, we use our \satnet\ model and train on \satlas\ images. For SeCo, we evaluate their original model trained on their dataset. We fine-tune the weights learned through self-supervision on the downstream tasks. We provide results for additional variants in supplementary Section B.3.

We fine-tune both the pre-training and self-supervised learning methods by first freezing the backbone and only training the prediction head for 32K examples, and then fine-tuning the entire model.
We provide additional experiment details in supplementary Section B.1.

\smallskip
\noindent
\textbf{Downstream Datasets.} The downstream tasks consist of four existing large-scale remote sensing datasets that involve classification with between 21 and 63 categories: UCM~\cite{ucmerced_land_use}, AID~\cite{aid}, RESISC45~\cite{resisc45}, and FMoW~\cite{fmow}. The other three are the Massachusetts Buildings and Massachusetts Roads datasets~\cite{massroad}, which involve semantic segmentation, and the Airbus Ships~\cite{airbus_ships} dataset, which involves instance segmentation.

\smallskip
\noindent
\textbf{Results.}
Table \ref{tab:downstream} shows downstream performance with varying training set sizes. \satlas\ consistently outperforms the baselines: when training on 50 examples, we improve average accuracy across the tasks by 18\% over ImageNet pre-training, and by 6\% over the next best baseline. The state-of-the-art performance achieved across such a wide range of downstream tasks clearly demonstrates the generalizability of the representations derived from \satlas\ pre-training, and the potential of \satlas\ to improve performance on the numerous niche remote sensing applications. We include results with more varying training examples in supplementary B.2.

\section{Use in AI-Generated Geospatial Data}

We have deployed \satlas\ to develop high-accuracy models for Satlas (\url{https://satlas.allen.ai/}), a platform for global geospatial data generated by AI from satellite imagery. Timely geospatial data, like the positions of wind turbines and solar farms, is critical for informing decisions in emissions reduction, disaster relief, urban planning, etc. However, high-quality global geospatial data products can be hard to find because manual curation is often cost-prohibitive. Satlas instead applies models fine-tuned for tasks like wind turbine detection to automatically extract geospatial data from satellite imagery on a monthly basis. Satlas currently consists of four geospatial data products: wind turbines, solar farms, offshore platforms, and tree cover.

\section{Conclusion} \label{sec:conclude}

By improving on existing datasets in both scale and label diversity, \satlas\ serves as an effective very-large-scale dataset for remote sensing methods. Pre-training on \satlas\ increases average downstream accuracy by 18\% over ImageNet and 6\% over existing remote sensing datasets, indicating that it can readily be applied to the long tail of remote sensing tasks that have few labeled examples.
We have already leveraged models pre-trained on \satlas\ to accurately detect wind turbines, solar farms, offshore platforms, and tree cover in the Satlas platform at \url{https://satlas.allen.ai/}.

\appendix

\section{Supplementary Material}

The supplementary material can be accessed from \url{https://github.com/allenai/satlas/blob/main/SatlasPretrain.md}.

{\small
\bibliographystyle{ieee_fullname}
\bibliography{egbib}
}

\end{document}